\documentclass[]{article}
\usepackage{proceed2e}
\usepackage{times}
\usepackage{tikz}
\usepackage{natbib}
\usepackage{amsmath}
\usepackage{amssymb}
\usepackage{amsthm}
\newtheorem{theorem}{Theorem}
\usepackage{algorithm}
\usepackage{algorithmic}
\usepackage{comment}
\usepackage{float}
\usepackage{hyperref}
\newcommand{\BEAS}{\begin{eqnarray*}}
\newcommand{\EEAS}{\end{eqnarray*}}
\newcommand{\BEQ}{\begin{equation}}
\newcommand{\EEQ}{\end{equation}}
\newcommand{\BIT}{\begin{itemize}}
\newcommand{\EIT}{\end{itemize}}

\newcommand{\reals}{{\mbox{\bf R}}}

\newcommand{\argmax}{\mathop{\rm argmax}}

\newcommand{\BB}{\mathbf{B}}
\newcommand{\DD}{\mathbf{D}}

\newcommand{\vv}{\mathbf{v}}

\newcommand{\xx}{\mathbf{x}}

\newcommand{\XX}{\mathbf{X}}
\newcommand{\WW}{\mathbf{W}}

\newcommand{\mmu}{\boldsymbol{\mu}}

\newcommand{\ttheta}{\boldsymbol{\theta}}

\newcommand{\yy}{\mathbf{y}}

\newcommand{\bb}{\mathbf{b}}

\newcommand{\pp}{\mathbf{p}}
\newcommand{\PP}{\mathbf{P}}
\newcommand{\qq}{\mathbf{q}}

\newcommand{\mL}{\mathcal{L}}

\newcounter{oursection}

\newcommand{\partd}[2]{\frac{\partial #1}{\partial #2}}

\usepackage{xspace}
\newcommand{\ds}{degrees of freedom }
\newcommand{\dso}{degrees of freedom. }
\newcommand{\Ds}{Degrees of freedom }
\newcommand{\df}{\textrm{df}}
\newcommand\iid{i.i.d.}

\title{Degrees of Freedom in Deep Neural Networks}

\author{ {\bf Tianxiang Gao} \and {\bf Vladimir Jojic} \\
	Computer Science Department\\
	University of North Carolina at Chapel Hill\\
	Chapel Hill, NC, 27514, USA\\
	\textit {\{tgao,vjojic\}@cs.unc.edu}}

\begin{document}
	
	\maketitle
	
	\begin{abstract}
		In this paper, we explore degrees of freedom in deep sigmoidal neural networks. We show that the degrees of freedom in these models are related to the expected \textit{optimism}, which is the expected difference between test error and training error. We provide an efficient Monte-Carlo method to estimate the degrees of freedom for multi-class classification methods. We show that the degrees of freedom is less than the parameter count in a simple XOR network. We extend these results to neural nets trained on synthetic and real data and investigate the impact of network's architecture and different regularization choices.
		The degrees of freedom in deep networks is dramatically less than the number of parameters. In some real datasets, the number of parameters is several orders of magnitude larger than the degrees of freedom. Further, we observe that for fixed number of parameters, deeper networks have less degrees of freedom exhibiting a regularization-by-depth. Finally, we show that the degrees of freedom of deep neural networks can be used in a model selection criterion. This criterion has comparable performance to cross-validation with lower computational cost.
	\end{abstract}
	
	\section{INTRODUCTION}
	\label{intro}
	
	Model selection is one of the key tasks in machine learning, as method's performance on training data is an optimistic estimate of its general performance. \citet{efron2004estimation} provided an estimate of optimism, difference of error on test and training data, and related it to a measure of model's complexity deemed effective degrees of freedom. This result reflects Occam's razor since models with higher degrees of freedom tends to have higher optimism. Degrees of freedom, defined as parameter counts, have been frequently used in model selection. However, even in linear models, the number of parameters are not a good indicator of model's complexity. Straightforward examples of this behavior are models fit using sparsity penalties. In that context, degrees of freedom are related to the number of non-zero parameters instead of total parameter count.
	
	\citet{ye1998measuring} introduced the concept of Generalized \Ds (GDF) for complex modeling procedures with Gaussian distributed outputs. GDF is defined based on the sensitivity of the fitted values to the perturbations in observed values. \citet{efron2004estimation} provided a framework for estimating degrees of freedom for modeling procedures with output in exponential family distribution. 
	In order to estimate \ds in deep neural networks for classification problems, where the outputs can be regarded as a categorical distribution, we extend Efron's results to the context of multinomial logistic regression. Similar to Ye's GDF, the computation of the degrees of freedom involves assessing network's changes in output as a result of perturbation of the training data. The more sensitive the network's output to the perturbation, the more degrees of freedom it has.
	
	We provide a straightforward algorithm for evaluating the degrees of freedom for any modeling procedure with outputs in categorical distribution form. This algorithm requires an additional run of the modeling procedure on the perturbed data. In the worst case, this amounts to doubling the running time of the procedure. Using this algorithm we first analyze the complexity of XOR network. This simple example highlights the fact that the degrees of freedom in a neural net is not simply equal to the total number of parameters in the network.
	
	In our experiments, we aim to answer following questions:
	\begin{enumerate}
		\item How does the network's complexity (DoF) vary with its architecture? Specifically, how do the degrees of freedom grow with the depth of a neural network?
		\item How does regularization affect network's complexity? Specifically, what is the impact of dropout, weight decay, adding noise on the degrees of freedom?
	\end{enumerate}
	
	We answer these questions in the context of feed-forward sigmoidal networks employed on classification tasks on both synthetic and real datasets.
	
	The prior work on the model complexity is rich, and we briefly review some key contributions.
	Bayesian Information Criterion (BIC) \citep{schwarz1978estimating} and Akaike Information Criterions (AIC) \citep{akaike1974new} are most commonly used techniques for model selection. Both aim to construct an estimate of the test log-likelihood by correcting the training set log likelihood with terms dependent on the number of parameters in the model in order to produce a score that is a less biased estimate of test log-likelihood. The weighting of the parameter count is different, BIC depends on the sample size, and AIC uses a constant. BIC applied to the family of models that contain the true model is consistent in the limit of the data. AIC, with some mild constraints, guarantees the selection of model with least square error, among models that do not include the true model. Crucial to the practical application of these methods is the correct count of parameters.
	Bayesian model selection elegantly avoids the need to specify the complexity of the network by evaluating evidence, a marginal probability of the data given the model. This approach marginalizes over all of the parameters, making models of different parameterizations comparable. The size of the parameter space directly impacts the evidence through this integration, as the prior on parameters gets spread thinly across high dimensional spaces. Unfortunately, the cost of computing such integrals is often prohibitive, but the models selected using these techniques have been shown to be very competitive. \citep{mackay2003information, neal2012bayesian, NIPSFeatureComp2003}.
	Kolmogorov-Chaitin complexity \citep{kolmogorov1965three} describes dataset complexity in terms of a program that recapitulates the data. Generation of task-specific neural networks using algorithmically simple programs was explored by \citet{schmidhuber1997discovering}. Networks whose parameters could not be captured by a simple program were avoided. A related method of Minimal Description Length reflects the desire for compact representation of the data. Its application \citep{HintonZemel2001} shows how the trade-off between the data and parameter compression can lead to an objective for training auto-encoders.
	Degrees of freedom of linear models fit with Lasso-type penalties have been analyzed, e.g. Lasso \citep{zou2007degrees}, Fused Lasso \citep{fusedlasso} and Group Lasso \citep{grouplasso}. The number of predictors and the number of degrees of freedom greatly differ due to the imposed sparsity and weight tying.
	Recent results on degrees of freedom for non-continuous procedures such as best subset regression and forward stagewise regression \citep{janson2015effective} highlight challenges in determining the complexity of these procedures as the estimators can be discontinuous.
	Research on Stein's Unbiased Risk Estimate has yielded model selection techniques \citep{stein1981estimation} as well as algorithms for their estimation \citep{ye1998measuring,ramani2008monte}. Generalization of SURE to exponential families has been proposed by \citet{eldar2009generalized}. However, its focus is on estimating parameter risk instead of prediction error. In linear models, the two neatly coincide. But this does not carry over to logistic regression and more broadly sigmoidal neural networks.
	
	\section{DEGREES OF FREEDOM FOR CATEGORICAL DISTRIBUTION}
	
	In this section, we derive the definition of \ds for categorical distribution from the optimism according to \cite{efron2004estimation}. Then, we introduce an efficient Monte-Carlo sampling based method \citep{ramani2008monte} to estimate degrees of freedom.
	
	\subsection{DEFINITIONS}
	
	We focus on models aimed at multi-class classification task. The data is assumed to be composed of features $\XX \in \mathbb{R}^{n \times p}$, and output labels $\yy$ range over $k$ categories. We will denote categorical distribution with $\mathcal{C}(\cdot)$. Categorical distribution over $k$ categories can be parameterized using a vector of non-negative values with a sum of 1. We treat sample label $y_i$ as realization of categorical random variables for a specific parameter vector $\mmu_i$. Hence $y_i \sim \mathcal{C}(\mmu_i)$, where $\mmu_i = [\mu_{i1},\mu_{i2},\dots,\mu_{ik}]$ is the \textbf{true probability} of sample $y_i$ being in each class. $\mu_{ic} \in [0,1]$ and $\sum_{c=1}^k \mu_{ic} = 1$.
	Members of exponential family follow form:
	\[
	f(\pp_i|\mmu_i) = r(\pp_i)\exp\{\ttheta(\mmu_i)^T \pp_i  -  A(\mmu_i)\}
	\]
	where $\pp_i$ is the vector of sufficient statistics for sample $i$, $\ttheta(\mmu_i)$ is the vector of natural parameters, $r(\pp_i)$ is the base measure, $A(\mmu_i)$ is the log-partition function. 
	
	For categorical distribution with parameter $\mmu_i$, we have $\pp_i = h(y_i) = [ \delta(y_i-1), \dots, \delta(y_i-k-1) ]^T$,
	where $\delta(\cdot)$ is the Kronecker delta function, $\delta(a)=1$ if $a=0$, $\delta(a)=0$ if $a \neq 0$. In other words, $\pp_i$ is a vector of the \textbf{observations} of sample $i$ being in each class. Base measure is $r(\pp_i) = 1$; natural parameters are $\theta_c(\mmu_i) = \ln \mu_{ic} - \ln (1 - \sum_{l=1}^{k-1} \mu_{il})$, and log partition function $A(\mmu_i) = \ln (1 + \sum_{c=1}^{k-1} e^{\theta_c(\mmu_i)})$.
	Note that both $\mmu_i$ and $\pp_i$ are of dimension $k-1$. Let $\PP = [ \pp_1, \dots, \pp_n ]^T$ be the matrix of observations for all sample labels $y_1,\dots,y_n$.
	
	\subsection{OPTIMISM IN MODELS WITH CATEGORICAL DISTRIBUTION}
	
	Optimism is the difference between expected test log deviance error and training log deviance error for a model fitting procedure. It is related to the complexity of the model and degrees of freedom is derived from optimism. If the optimism for a modeling procedure can be estimated, we can use it for model selection. \cite{efron2004estimation} provides the derivations of expected optimism for the single parameter exponential family. We follow Efron's approach to derive the definition of \ds for modeling procedure with output in categorical distribution form.
	
	Given sample input $\xx_i$, we assume that the output label $y_i \sim \mathcal{C}(\mmu_i)$. Let $\hat{\mmu}_i = \mL(\pp_i)$ be the \textbf{estimated probability} for sample $i$ from observations $\pp_i$. The log deviance error for $\hat{\mmu}_i$ and $\pp_i$ is:
	\BEAS
	\textrm{err}_i & = & -2 \log f(\pp_i|\hat{\mmu}_i^T) \\
	& = & -2 [ \ttheta(\hat{\mmu}_i)^T\pp_i - A(\hat{\mmu}_i) ]
	\EEAS
	
	Suppose we have another sample $y^0_i$ drawn from the same distribution as $y_i$, $y^0_i \sim \mathcal{C}(\mu_i)$. Let $\qq_i = h(y^0_i)$ be the vector of its observations. The expected log deviance error of $\qq_i$ using $\hat{\mmu}_i$ is:
	\BEAS
	\textrm{Err}_i & = & E_{y^0_i} \{ -2 \log f(\qq_i|\hat{\mmu}_i) \} \\
	& = & -2 [\ttheta(\hat{\mmu}_i)^T \mmu_i + A(\hat{\mmu}_i)]
	\EEAS
	
	The definition of optimism is:
	\BEAS
	O_i & = & \textrm{Err}_i - \textrm{err}_i \\
	& = & 2 \ttheta(\hat{\mmu}_i)^T(\pp_i - \mmu_i)
	\EEAS
	Hence, optimism is the difference between log deviance error on the training set and expected log deviance error with respect to the true distribution. 
	
	The expected optimism over $y_i \sim \mathcal{C}(\mmu_i)$ for the estimated probability $\hat{\mmu}_i$ and true probability $\mmu_i$ is:
	\BEAS
	\Omega_i & = & 2 E_{y_i}\{ \ttheta(\hat{\mmu}_i)^T(\pp_i - \mmu_i). \}
	\EEAS
	
	As we do not know the true probability $\mmu_i$, we cannot compute the expected optimism. However, we can get an approximate measurement using Taylor series expansion. We can approximate $\ttheta(\hat{\mmu}_i)$ by taking the Taylor series expansion at $\pp_i = \mmu_i$ to obtain:
	\BEAS
	\ttheta(\hat{\mmu}_i) & \approx & \ttheta(\mL(\mmu_i)) +  \DD^{(i)} (\pp_i-\mmu_i).\\
	\EEAS
	$\DD^{(i)}$ is the first derivative matrix where each entry $D^{(i)}_{jc} = \partd{\theta_j(\mL(\vv))}{v_{c}} |_{\vv = \mmu_i}$.
	
	Therefore, we can approximate expected optimism as:
	\BEAS
	\tilde{\Omega}_i  & = & 2 E_{y_i} \{ [\ttheta(\mL(\mmu_i)) +  \DD^{(i)} (\pp_i-\mmu_i) ]^T (\pp_i - \mmu_i) \}\\
	& = & 2  E_{y_i} \{ \sum_{j=1}^{k-1} \sum_{l=1}^{k-1} (p_{ij}-\mu_{ij})(p_{il} - \mu_{il}) D^{(i)}_{jl} \}\\
	& = & 2 \sum_{j=1}^{k-1} \sum_{l=1}^{k-1} \textrm{cov}(p_{ij},p_{il}) \partd{\theta_j(\mL(\vv))}{v_{l}} |_{\vv = \mmu_i} \\
	\EEAS
	We can estimate the expected optimism by assuming $p_i \sim \mathcal{C}(\hat{\mmu}_i)$, so:
	
	\begin{equation}
	\label{eqn:hat}
	\hat{\Omega}_i = 2 \sum_{j=1}^{k-1} \sum_{l=1}^{k-1} \textrm{cov}(p_{ij},p_{il}) \partd{\theta_j(\mL(\vv))}{v_{l}} |_{\vv = \hat{\mmu}_i}.
	\end{equation}
	In categorical distribution, $\textrm{cov}(p_{ij},p_{il}) = - \hat{\mu}_{ij}\hat{\mu}_{il}$, if $i \neq j$. $\textrm{var}(p_{ij}) = \hat{\mu}_{ij}(1-\hat{\mu}_{ij})$. Therefore, Equation~(\ref{eqn:hat}) can be reduced to:
	
	\begin{equation}
	\label{central}
	\hat{\Omega}_i =  2\sum_{j=1}^{k-1} \partd{\mL_j(\vv)}{v_j} |_{\vv = \hat{\mmu}_i}. \\
	\end{equation}
	
	The proof is given in the supplementary material.
	
	Equation~(\ref{central}) for $k=2$ is exactly the result for Bernoulli distribution derived in \citep{efron2004estimation}. Efron also showed that Eqn~(\ref{central}) gives the correct degrees of freedom for maximum likelihood estimation \citep{efron1975defining}. In a $p$-parameter curved exponential family, we have:
	\[
	\sum_{i=1}^{n} \partd{\mL(\vv)}{v_{i}} |_{\vv = \hat{\mmu}_i}  = p.\\
	\]
	
	
	Here, we define the degrees of freedom for a classification model estimator $\hat{\mmu}_i = \mL_i(\PP)$ on all the data samples $\PP$ to be:
	\begin{equation}
	\label{df}
	\df = \sum_{i=1}^n \sum_{c=1}^{k-1} \partd{\mL_{ic}(\PP)}{p_{ic}}.
	\end{equation}
	
	This definition tells that the \ds is the sum of each sample's sensitivity of its estimated probability to the perturbations in its observation for all categories. 
	
	\subsection{DEGREES OF FREEDOM FOR MODEL SELECTION}
	
	As \ds is related to the expected optimism, we can use \ds for model selection. According to Equation~(\ref{central}) and~(\ref{df}), the relationship between expected test and training log deviance errors is:
	\begin{equation}
	\label{optaic}
	\sum_{i=1}^n E_{y_i}\{\textrm{Err}_i\} = \sum_{i=1}^n E_{y_i} \{\textrm{err}_i\} + 2 \textrm{df}.
	\end{equation}
	
	Euqation~(\ref{optaic}) is very similar to Akaike Information Criterions (AIC) \cite{akaike1974new}:
	\begin{equation}
	\label{aic}
	\textrm{AIC} = \sum_{i=1}^n \textrm{err}_i + 2 k,
	\end{equation}
	where $k$ is the number of parameters. We refer to $2\df$ in Equation~(\ref{optaic}) and $2k$ in Equation~(\ref{aic}) as ``complexity correction" for training log deviance error. In simple linear regression models, $\df = k$, and the complexity corrections are the same. However, in complex models such as deep neural networks, simply counting number of parameters can result in overestimate of the expected test log deviance error. Therefore, we introduce $\textrm{DoFAIC}$ for model selection:
	\begin{equation}
	\label{dofaic}
	\textrm{DoFAIC} = \sum_{i=1}^n \textrm{err}_i + 2 \textrm{df}.
	\end{equation}
	
	DoFAIC uses \ds instead of the number of parameters for complexity correction. We assume that DoFAIC can produce a better criterion for model selection than Na\"{\i}ve AIC.
	
	\subsection{MONTE-CARLO ESTIMATE FOR DEGREES OF FREEDOM}
	
	For most practical estimators of the model's predictions with respect to the data derivatives, $\partd{\mL_{ic}(\PP)}{p_{ic}}$ are not available in closed form. For example, fitting multinomial logistic regression using stochastic gradient descent with adaptive learning rates requires a fairly sophisticated derivation which accounts for changes in step-sizes as a result of data perturbation. For deep neural networks, this difficulty grows due to the use of back-propagation. In this paper, we used a sampling based method to efficiently estimate 
	
	\paragraph{Monte-Carlo estimation}
	A theoretical result for a stochastic estimate of the degrees of freedom of nonlinear estimators has been proposed by   \citet{ramani2008monte}. We restate the key result from that paper here.
	\begin{theorem} Let $\bb$ be a zero mean \iid\ random vector (that is independent of y) with unit variance and bounded higher moments. Then
		\[
		\sum_i \frac{\partial f(\yy)}{\partial y_i} = \lim_{\epsilon \rightarrow 0} E_{\bb}\left[
		\bb^T\left(\frac{f(\yy+\epsilon\bb) - f(\yy)}{\epsilon}\right)\right]
		\]
		provided that $f$ admits a well-defined second-order Taylor expansion.
	\end{theorem}
	We sketch out a proof that the prediction in a neural net via forward pass is a smooth function of the observations of training labels. We will abbreviate ``differentiable with respect to observations'' as d.w.r.t.o.
	Sigmoid and soft-max are smooth functions of their inputs. The cross-entropy loss is a multivariate function that depends on data and weights, and all of its partial derivatives exist. For simplicity, we assume that the network is trained using gradient descent. Each update of the network's parameters is a linear combination of previous weights and a gradient of the loss. Assuming that the initial weights d.w.r.t.o. and loss is smooth then the update yields weights that are d.w.r.t.o. Random initialization and pre-training both yield initializations that are independent of observations, hence the partial derivatives of the initial weights with respect to observations are 0. By induction, gradient descent, at any iteration, yields weights that are d.w.r.t.o. Forward pass through sigmoidal network yields estimated probabilities which are smooth with respect to observations. Thus, the Taylor expansion required by the above theorem exists.
	
	Using this theorem, we can evaluate the derivative of a function $\partd{f(x)}{x}$ by perturbing the inputs. We applied a modified version of the method \citep{ramani2008monte} for categorical distribution. We applied random perturbation to the observations to estimate the degrees of freedom:
	\BEAS
	& \df =  \sum_{i=1}^n \sum_{c=1}^{k-1} \partd{\mL_{ic}(\PP)}{p_{ic}} \\
	& =  \lim_{\epsilon \to 0} \left\{ E_{\BB} \left[ \sum_{i}\sum_{c} b_{ic} \left( \frac{\mL_{ic}(\PP + \epsilon \BB) - \mL_{ic}(\PP)}{\epsilon}\right)\right] \right\},
	\EEAS
	where $\BB$ is a zero-mean \iid\ random matrix with unit variance and bounded higher order moments. Therefore, we can approximate $df$ with $T$ independent samplings of $\BB^{(t)}$:
	\begin{equation}
	\label{mc}
	df \approx \frac{1}{T}\sum_{t=1}^T \sum_{i=1}^n \sum_{c=1}^{k-1} b^{(t)}_{ic} \left( \frac{\mL_{ic}(\PP + \epsilon \BB^{(t)}) - \mL_{ic}(\PP)}{\epsilon}\right),
	\end{equation}
	where $\epsilon$ is a small value. In our experiments, we choose $\epsilon = 10^{-5}$. To better estimate the sensitivity, we can use the average of multiple runs as the final estimation. The algorithm for estimating \ds is summarized in Algorithm~\ref{(alg)}.
	
	\renewcommand\algorithmicrequire{\textbf{Input:}}
	\begin{algorithm}[t!]
		\begin{algorithmic}[1]
			\REQUIRE training data $\XX \in \reals^{n\times p} $, $\yy \in \{1,2,..,k\}^n$\\
			\STATE    Compute observations matrix $\PP = h(\yy)$\;
			\STATE    Train model on $\XX$ and $\PP$
			\STATE    Compute estimated probabilities for each sample $\mL(\PP)$\;
			\STATE    Sample entries of $\BB^{(t)} \in \reals^{p \times k}$ from zero-mean, unit variance normal distribution\;
			\STATE    Train model on $\XX$ and $\PP^{(t)} = \PP + \epsilon \BB^{(t)}$\;
			\STATE    Using trained model compute estimated probabilities for each sample $\mL(\PP^{(t)})$;
			\STATE      Repeat 4-6 for $T$ times;
			\STATE    Calculate $df$ from Equation (\ref{mc})
		\end{algorithmic}
		\caption{Monte Carlo algorithm for computing degrees of freedom of a multi-class classifier \label{(alg)}}
	\end{algorithm}
	
	Note that training on original and perturbed observations matrix can be performed in parallel. Finally, we also derived analytical derivatives for stochastic gradient descent learning which yields the same degrees of freedom as the algorithm presented above. However, this method requires maintenance of partial derivatives of each parameter with respect to each sample's observations. Such storage requirements make this method impractical for real world applications.
	
	\paragraph{Variance reduction} For deep neural networks, training takes a considerable amount of time. In order to estimate \ds in a reasonable computational time, we used a variance reduction technique -- common random numbers -- during Monte-Carlo sampling. When comparing the \ds on a specific data, fixed $\PP$, for several different fitting procedures, we used the same perturbation matrix $\BB$ for all the models. We used the same random seed for all models throughout the training. For example, in deep neural network training, we use the same random seed to initialize weights and bias; during pre-training with denoising-autoencoders, we use the same random seed for drop-out and input corruptions. For stochastic gradient descent methods, we use the same mini-batches splittings during training. In our experiment, we found that we can estimate \ds well enough using just one perturbed copy of the data when using these variance reduction techniques. 
	
	\subsection{DEGREES OF FREEDOM IN MULTINOMIAL LOGISTIC REGRESSION}
	
	In order to validate the above algorithm in a setting with known degrees of freedom, we perform an empirical analysis of the \ds in different multinomial logistic regression models.
	
	We generate an \iid\ zero mean unit variance random design matrix $\XX$ with $n = 100$ samples and $p = 20$ features. We represent each sample with $\xx_i = [x_{i1}, x_{i2}, \dots, x_{ip}]$. With $k=4$ class, we generated a random weight matrix $\WW \in \mathbb{R}^{p \times k}$, where each entry $w_{ic} \sim \mathcal{N}(0,1)$. We generate each label from $y_i = \argmax_j \mu_{ij}$, where $\mmu_i = e^{\xx_i \WW}$.
	
	We fit 5 models using multinomial logistic regression. In $i$th model, we only use first $2i$ features in $\XX$ to fit. Therefore, $i$th model only contains $2(i+1)(k-1)$ parameters and the degrees of freedom are equal to the number of parameters. We perform 5 Monte-Carlo \ds estimates for each model.
	
	\begin{figure}[h]
		\centerline{\includegraphics[width=9.4cm]{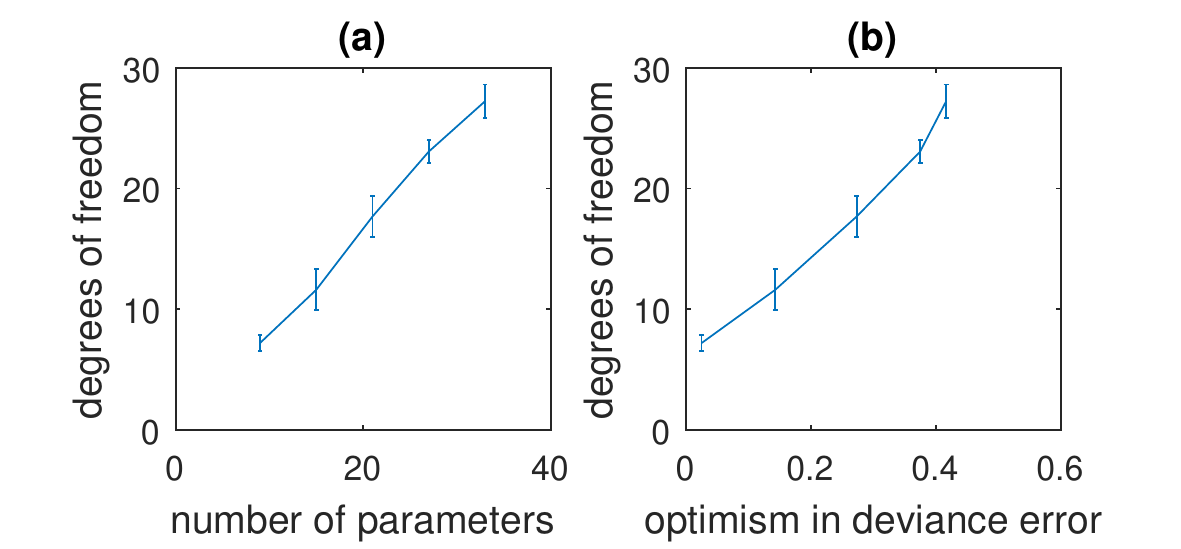}}
		\caption{\textbf{(a)} Comparison between \ds estimates in multinomial logsitic regression and the true number of parameters used in the model. \textbf{(a)} Comparison between \ds estimates in multinomial logsitic regression and the optimism in log deviance error. In each plot, blue line is the mean of the five Monte-Carlo estimates. Error bar represents the standard error.\label{fig:mnr}}
	\end{figure}

	We plot \ds in Figure~\ref{fig:mnr}(a). We observed that \ds are very close to the number of parameters we used in the model. The standard error for Monte-Carlo estimate is small.
	
	We also randomly generated 1000 samples for testing. Optimism is calculated by the difference between average testing log deviance error and training log deviance error. We plot the \ds and optimisms for all 5 models in Figure~\ref{fig:mnr}(b). It shows that the optimism has a linear relationship with degrees of freedom, as expected.
	
	\subsection{DEGREES OF FREEDOM OF A XOR NETWORK}
	
\begin{figure}[h]
		\begin{center}
			\begin{tikzpicture}[minimum size=0.5cm,->,shorten >=1pt,auto,node distance=0.5cm,semithick]
			\tikzstyle{var}=[circle,fill=white,draw=black,text=black]
			\tikzstyle{varh}=[circle,fill=red,draw=black,text=black]
			\tikzstyle{vare}=[circle,fill=green,draw=black,text=black]
			\tikzstyle{vard}=[circle,fill=blue,draw=black,text=black]
			\node[var] (X1) at (-2,2)  {$x_1$};
			\node[var] (X2) at (-2,0)  {$x_2$};
			\node[var] (H1) at (0,2)  {$h_1$};
			\node[var] (H2) at (0,0)  {$h_2$};
			\node[var] (Y) at (2,1)  {$o$};
			\path[line width = 1] (X1) edge node {} (H1);
			\path[line width = 1] (X1) edge node {} (H2);
			\path[line width = 1] (X2) edge node {} (H1);
			\path[line width = 1] (X2) edge node {} (H2);
			\path[line width = 1] (H1) edge node {} (Y);
			\path[line width = 1] (H2) edge node {} (Y);
			\end{tikzpicture}
			\caption{A Neural Network with 2 Hidden Nodes \label{nn}}
		\end{center}
	\end{figure}
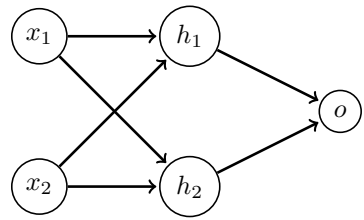

	We generated a small synthetic example using exclusive-or (XOR) operator, where $\textrm{XOR}(a,b) = 0$ if $a=b$, and $\textrm{XOR}(a,b)=1$ if $a \neq b$. Given an input $x_1,x_2 \in \{0,1\}$, the output $y = XOR(x_1,x_2)$, we hope to learn a model of XOR operator. In general, we can build a neural network with two hidden nodes as shown in Figure~\ref{nn} and weights in Table~\ref{XOR} to learn a perfect XOR classifier.
	
	\begin{table}[h]
		\caption{An XOR Network}
		\label{XOR}
		\begin{center}
			\begin{tabular}{|l|c|c|c|r|}
				\hline
				$x_1$ & 0 & 1 & 0 & 1 \\ \hline
				$x_2$ & 0 & 0 & 1 & 1 \\ \hline
				$h_1 = \sigma(F(-0.5 + x_1 - x_2))$ & 0 & 1 & 0 & 0 \\ \hline
				$h_2 = \sigma(F(-0.5 - x_1 + x_2))$ & 0 & 0 & 1 & 0 \\ \hline
				$y = \sigma(K(-0.5 + h_1 + h_2))$ & 0 & 1 & 1 & 0 \\ \hline
			\end{tabular}
		\end{center}
	\end{table}
	
	A network that trained properly should have weight matrix with form in Table~\ref{XOR}. If $x$ contains no noise, $F$, a multiplier, can be infinitely large to achieve perfect estimation. Therefore, we set $y$ to be $0.9$ instead of 1. 
	
	We train networks with different structures on XOR data using back-propagation and estimate their \ds using Monte-Carlo method. Even though there are 9 parameters in the network, we found that the \ds for all learned models is 4. We note that the symmetry in weights of the inputs to the two hidden nodes, eliminates degrees of freedom, as does implicit tying of the weights of inputs to the output node. To give an intuition why this tying occurs, we note that the predominantly correctly labeled data drives the network to keep the weights close to each other. Hence, a small perturbation in the labels can affect multiple weights simultaneously, but does not disturb their balance. This observation encourages us to investigate deeper models.

	\section{DEGREES OF FREEDOM IN DEEP NEURAL NETWORKS}
	
	In this section, we investigate \ds in deep neural network models. From the XOR example, we know that the degrees of freedom in a network is not equal the number of parameters in the model. The structure of the network and different regularization techniques will impact \dso
	
	\subsection{TERMINOLOGIES AND SETTINGS}
	
	In the following experiments, we explore deep networks trained to solve larger classification problems. Each of the networks takes real value vector $\xx_i \in \mathbb{R}^{p \times 1}$ as input and outputs the probability $\hat{\mmu}_i$ for this sample being in one of $k$ categories. We use sigmoid activation function for all the hidden nodes and a soft-max in the last layer. The number of hidden layers is called ``\textbf{depth}'' of the network. We only consider networks with an equal number of units in each hidden layer, and we call this number ``\textbf{width}'' of the network. Next, we investigate \ds in networks with different width and depth.
	
	\paragraph{Stacked-Auto-Encoder (SdA) pre-training} We used SdA \citep{vincent2010stacked} to pre-train the neural network with input dataset, as unsupervised pre-training helps the network to achieve a better generalization from the training data on supervised tasks \citep{erhan2010does}. In denoising auto-encoder, \textbf{corruption} is used in layer-wised pre-training. The corruption is introduced by zeroing out input to the auto-encoder with a certain probability. The chosen probability of corruption is called \textbf{corruption rate}. \textbf{Dropout} \citep{srivastava2014dropout} is also used during the pre-training of SdA, where output of hidden units are randomly zeroed with probability, which is called \textbf{dropout rate}. We assume that increasing in corruption rate or dropout rate will reduce \ds as they provide more regularization to the network.
	
	\paragraph{Weight-decay} We used a weight decay penalty on the sum of the squares of all the weights in the network during both pre-training and fine-tuning stage. Adding this penalty prevents the network from over-fitting. We refer to the multiplier associated with the sum of squares as \textbf{weight decay rate}. We expect to see that the \ds drops with increasing weight decay rate.
	
	\paragraph{Implementation} All our code are based on Theano \citep{Bastien-Theano-2012,bergstra+al:2010-scipy} and we ran experiments on a cluster of machines with NVIDIA Tesla compute cards.
	
	\subsection{DATA SETS}
	
	We prepared a synthetic dataset and two real datasets MNIST and CIFAR-10 to estimate \dso
	
	\paragraph{Synthetic} We build a synthetic dataset from a randomly generated network with 30 input nodes, 2 hidden layers with 30 hidden nodes in each, and 4 output nodes. We generated $n=5000$ random zero-mean unit variance inputs with 30 dimensions. Each layer was fully connected to the previous layer, and we generated weights $w \sim \mathcal{N}(0,5)$. We used sigmoid activation function for each layer and a soft-max on top of the network. The output sample labels $y$ are then sampled according to the probabilities from the soft-max layer. To get the optimism, we also generated another 5000 samples for test.
	
	\paragraph{MNIST} \footnote{\url{http://yann.lecun.com/expdb/mnist/}} \citep{lecun1998mnist} is a benchmark dataset that contains handwritten digit images. Each sample is a $28 \times 28$ image from 10 classes. We used 50000 samples for training.
	
	\paragraph{CIFAR-10} \footnote{\url{https://www.cs.toronto.edu/~kriz/cifar.html}}
	\citep{krizhevsky2009learning} is a dataset contains $32 \times 32$ tiny color images from 10 classes. Each sample has $3072$ features. We used 50000 samples for training.
	
	\subsection{DEGREES OF FREEDOM AND THE STRUCTURE OF THE NETWORK}
	\label{sec:struct}
\begin{figure}[h]
		\centerline{\includegraphics[width=9cm]{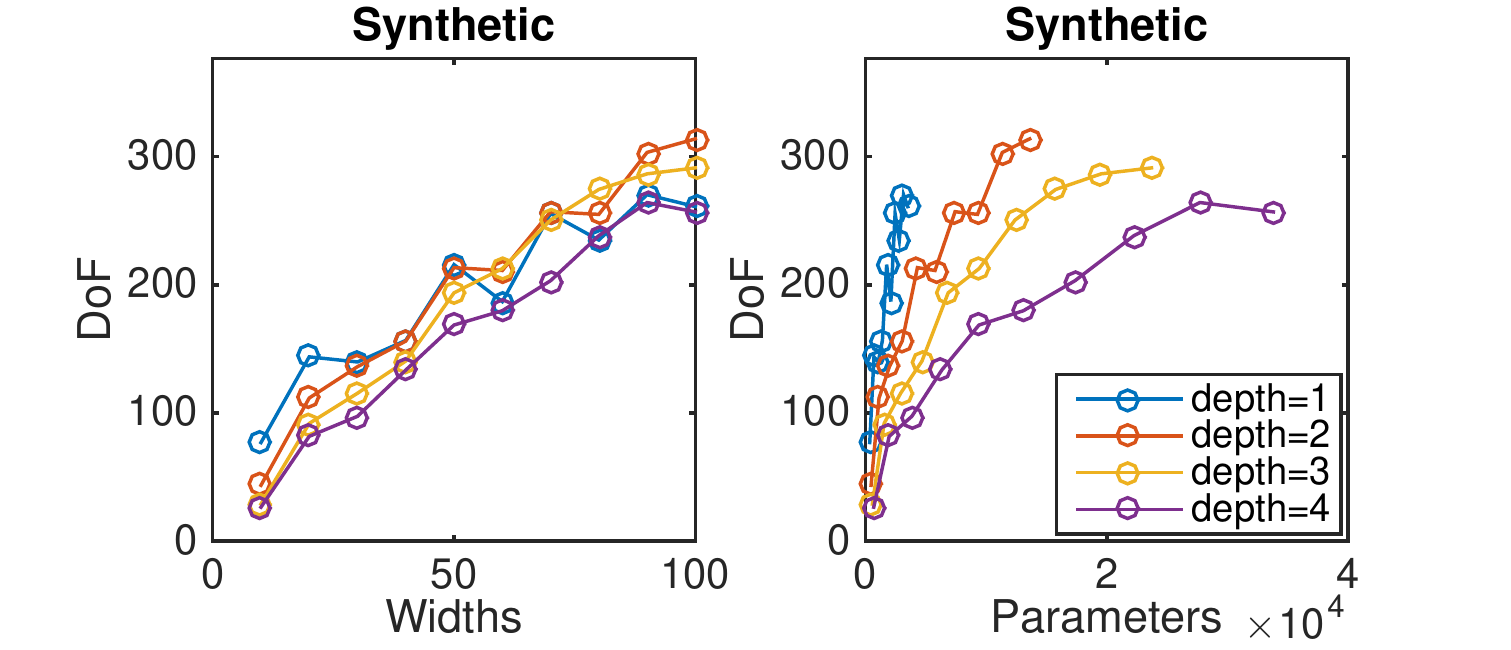}}
		\caption{\Ds estimates for different models trained on synthetic data. Left: \ds vs network width. Right: \ds vs number of parameters in the network, which is linearly related to the network depth and quadratically related to the number of width. The lines represent the \ds estimate from 1 Monte-Carlo run, and the color of each indicates the depth of the models.\label{fig:ss}}
	\end{figure}

	To investigate the \ds for networks with different structures, we estimated the \ds for networks with width $[10,20,\dots,100]$ and depth with 1,2,3 and 4, where all the hidden layers have equal widths. We used SdA to pre-train with 0.1 dropout rate and 0.1 corruption rate. We use weight decay penalty $1e-5$ for both pre-training and fine-tuning. The estimated \ds is shown in Figure~\ref{fig:ss}.
	
	From the results, we found that networks with more width have more \dso This is reasonable as increasing width leads to more independence between parameters. However, the \ds in deep networks is generally much less than the number of parameters it used. We see that the ratio of the parameters to degrees of freedom is on the order of $10^2$. Loosely, one degree of freedom is acquired for 100 parameters. Among the models with the same number of parameters, deeper networks have less \dso This observation indicates that the depth of the network has regularization on the complexity. 
	
\begin{figure}[h]
		\centerline{\includegraphics[width=9cm]{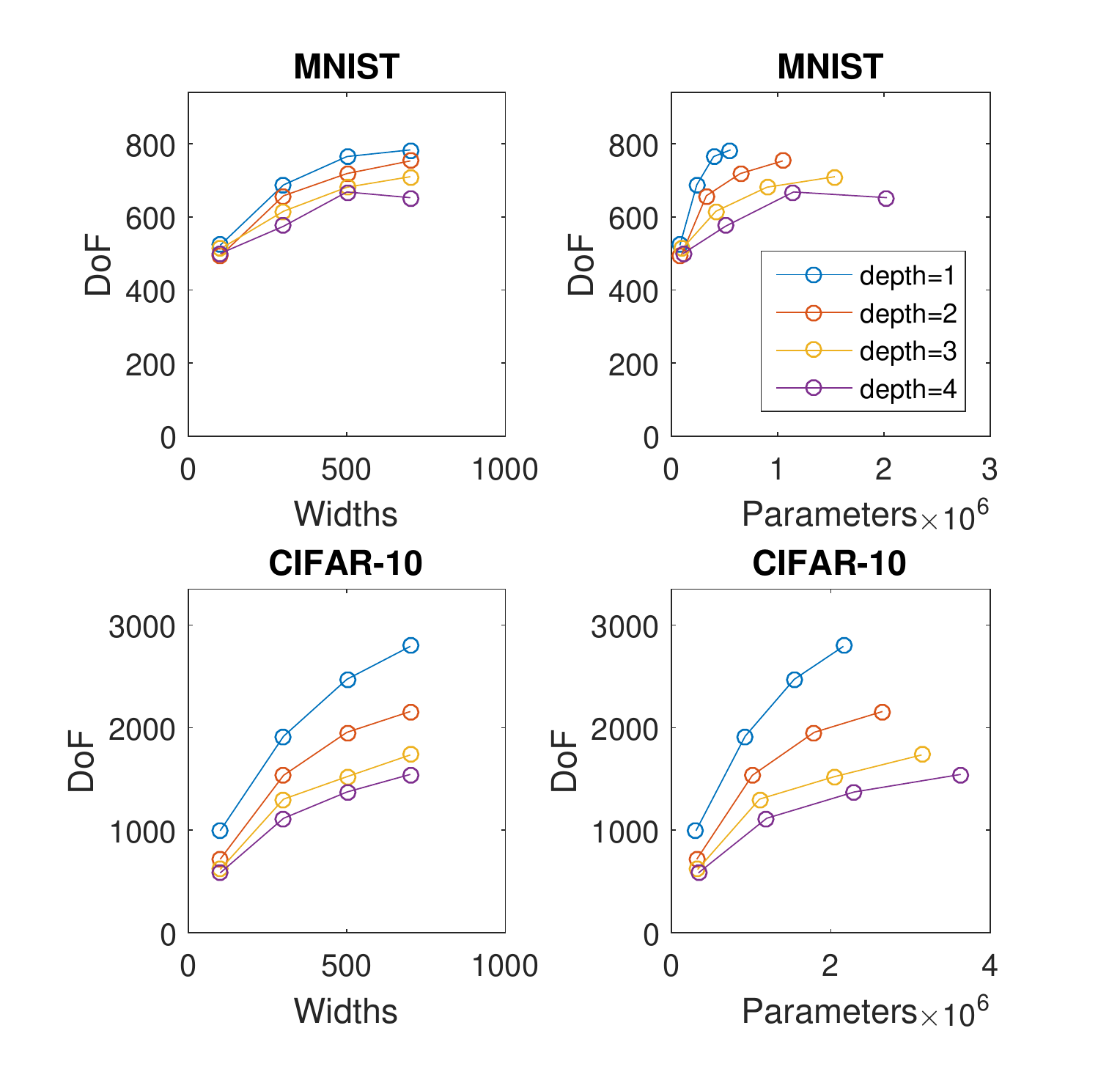}}
		\caption{\Ds estimates for different models trained on MNIST and CIFAR-10. Left: \ds vs network width. Right: \ds vs the number of parameters in the network, which is linearly related to the network depth and quadratically related to the number of widths. The lines represent the \ds estimate from single Monte-Carlo sample and the color of each indicates the depth of that model.\label{fig:rs}}
	\end{figure}

	To further validate our assumption that deeper networks have less degrees of freedom, we also estimated \ds on MNIST and CIFAR-10 dataset. We tested networks with width $[100,300,500,700]$, all other settings are the same as in the above synthetic experiment. The results are shown in Figure~\ref{fig:rs}.
	
	We observe that we can make the same conclusions hold for MNIST and CIFAR-10 as we did for synthetic data. The only difference is increasing depth results in more \ds than models trained with synthetic data. We attribute this to the differences of input data size and complexity between the real datasets, MNIST and CIFAR-10, and the much simpler synthetic datasets.
	
	\subsection{DEGREES OF FREEDOM AND REGULARIZATION TECHNIQUES}
	
	When training a deep neural network, many practical methods can be used for regularization. We investigate how the different techniques affect the \ds in the model.
	
	We train networks using the same settings as in Section~\ref{sec:struct}. In this experiment, we separately trained networks with different settings of penalty rates: corruption rate, dropout rate, and weight decay rate. We changed one rate at a time while keeping rest fixed.

	tested the corruption rate, dropout rate, and weight decay penalty by keeping all others fixed and only changing one at a time.
	
\begin{figure}[h]
		\centerline{\includegraphics[width=8cm]{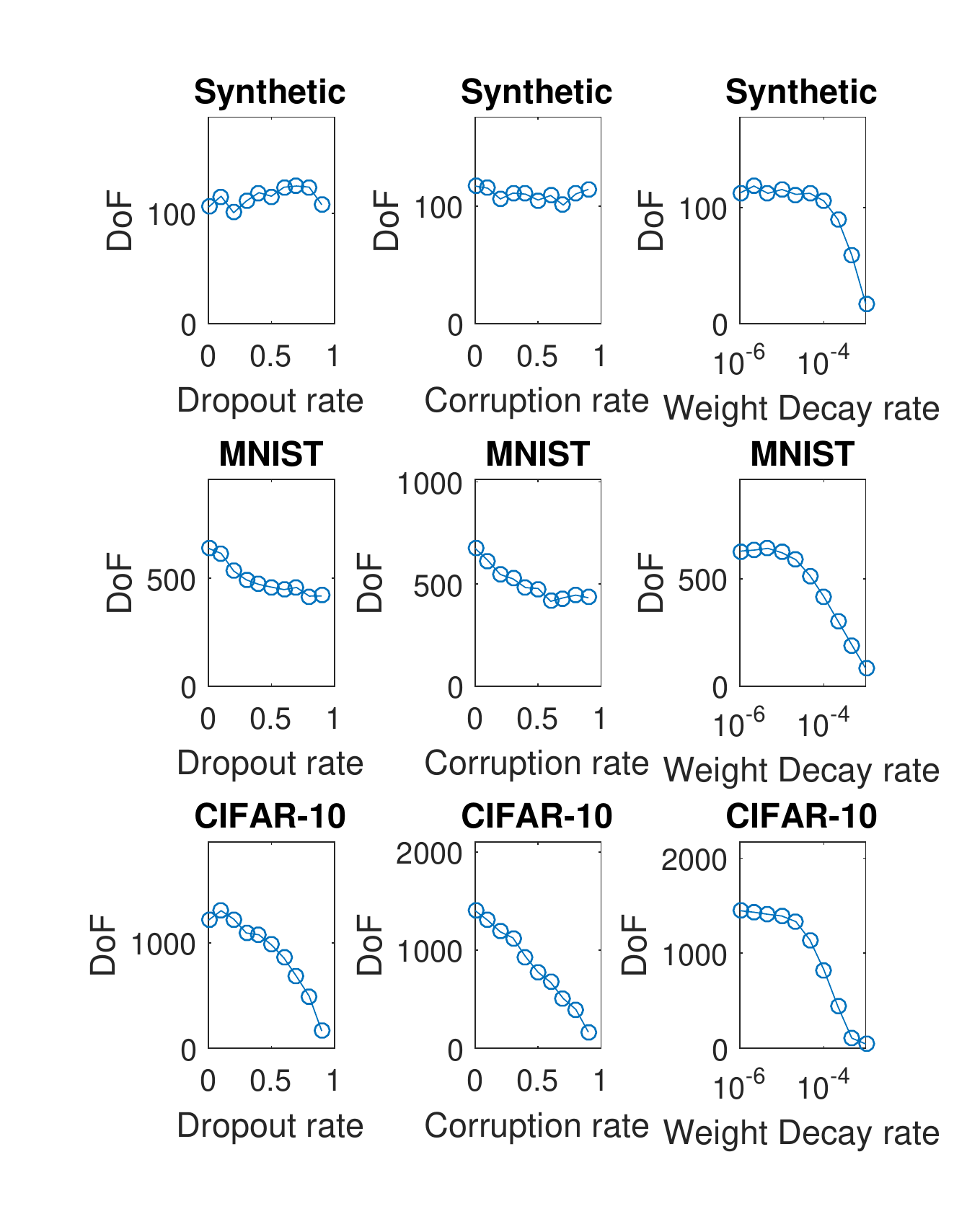}}
		\caption{\Ds estimates for models trained on Synthetic data, MNIST and CIFAR-10 under different regularizations\label{fig:regs}. The lines represent the \ds estimate.}
	\end{figure}

	For all three datasets, we trained network using corruption rate and dropout rate from $[0,0.1,0.2,\dots,0.9]$, and weight decay rate from $10^{-6}$ to $10^{-3}$. For each setting of regularization parameters, we trained a 3 layer network $[30,30,30]$ for synthetic data and $[300,300,300]$ on MNIST and CIFAR-10 data. We used one Monte Carlo sample to estimate \ds in each model. The result is shown in Figure~\ref{fig:regs}.

	We found that neither corruption rate nor dropout rate affected \ds drastically for synthetic data. This is because the input of the synthetic data is generated randomly. Hence, pre-training cannot learn higher level features for synthetic data. For MNIST and CIFAR-10, we found that both corruption rate and dropout rate have an impact on \ds. In CIFAR-10, the regularization effect is much larger. These results suggest that the regularization strength from dropout and corruption can be data-specific. 
	
	Weight decay penalty has a very strong effect on the \ds for all three datasets. Further, the weight decay exhibited a highly non-linear impact on the degrees of freedom, in dramatic contrast to its effect in ridge regression.\footnote{Ridge regression degrees of freedom scale with $\frac{1}{1+\lambda}$ which is non-linear but much tamer multiplier than in neural networks}
	
	\subsection{MODEL SELECTION USING DEGREES OF FREEDOM}
	
\begin{figure}[h]
		\centerline{\includegraphics[width=10cm]{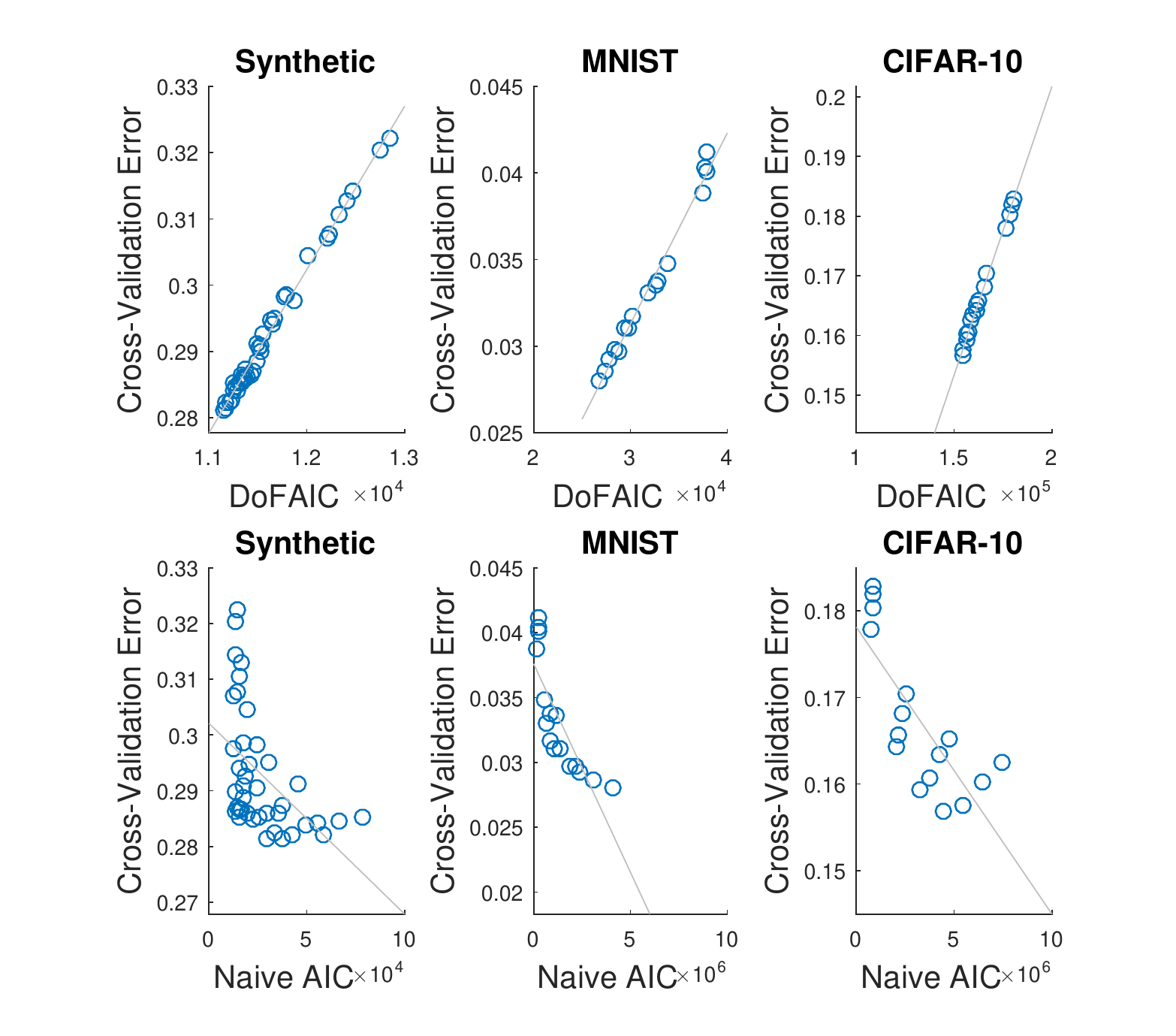}}
		\caption{\label{fig:cv} Comparison between DoFAIC (first row) / Na\"{\i}ve AIC (second row) and 5-fold cross validation. Each circle in the plot represents a model with a specific structure. The x-axis is the mean cross-validation log deviance error across 5 folds.}
	\end{figure}

	To validate that DoFAIC is a useful criterion for model selection, we compare it against model selection based on error estimates using cross validation. For brevity, we refer to the cross validation estimate of error as cross validation error. We performed a 5-fold cross-validation experiment for Synthetic, MNIST and CIFAR data on models with different network structures learned in Section~\ref{sec:struct}. We calculated DoFAICs for all the models we trained using Equation~(\ref{dofaic}) with the estimated degrees of freedom. We also calculated Na\"{\i}ve AIC using Equation~(\ref{aic}) with the number of parameters in the network. We compared these estimates against cross-validation errors. The result is shown in Figure~\ref{fig:cv}.
	
	Further, we calculate the Spearman rank correlation between cross-validation log deviance errors and DoFAIC/Na\"{\i}ve AIC estimates for each dataset. The result is shown in Table~\ref{opt}.        
	
	\begin{table}[h]
		\centering
		\caption{Spearman Rank Correlation between Cross-validation error and DoFAIC/Na\"{\i}ve AIC}
		\label{opt}
		\begin{tabular}{|l|l|l|}
			\hline
			Dataset   & DoFAIC $\rho$ & Na\"{\i}ve AIC $\rho$ \\ \hline
			Synthetic & 0.9865 & -0.6711\\ \hline
			MNIST     & 0.9853 & -0.9471 \\ \hline
			CIFAR-10  & 0.9941 & -0.7824 \\ \hline
		\end{tabular}
	\end{table}

	We find that DoFAIC is very consistent with cross-validation error. Na\"{\i}ve AIC, on the other hand, exhibits negative correlation with cross validation error due to highly non-linear behavior. This is because Na\"{\i}ve AIC overestimates the complexity of the model by using the large number of parameters in the network. The actual complexity in deeper and larger networks are much less than the number of parameters.

	For all three datasets, both DoFAIC and cross-validation chose the same model. This indicates that DoFAIC can be used for model selection. We note that $k$-fold cross-validation, which needs at most $k$ rounds of training, while DoFAIC only requires at most 2 rounds of training. This makes DoFAIC an efficient model selection criterion.
	
	\section{DISCUSSION}
	
	In this paper, we investigated the \ds for  classification models and presented an efficient method to estimate their degrees of freedom. We showed that for simple classification models, \ds is equal to the number of parameters in the model. In deep networks, the \ds is generally much less than the number of parameters in the model, and deeper networks tend to have less degrees of freedom. We also theoretically and empirically showed we can use DoFAIC as an efficient criterion for model selection, which has comparable performance to cross-validation.
	
	\paragraph{Future work} It would be interesting to investigate \ds in other deep architectures, such as Convolution Neural Network (CNN), Recurrent Neural Networks (RNN), denoising auto-encoders and contractive auto-encoders.
	
	\subsubsection*{Acknowledgement} This work was supported by NSF INSPIRE award IOS-1343020 to VJ.
	
	\bibliography{DNNDF}
	\bibliographystyle{plainnat}
\end{document}